\documentclass[lettersize, journal]{IEEEtran}
\usepackage{amsmath,amsfonts}
\usepackage{algorithmic}
\usepackage{array}
\usepackage[caption=false,font=normalsize,labelfont=sf,textfont=sf]{subfig}
\usepackage{textcomp}
\usepackage{stfloats}
\usepackage{url}
\usepackage{verbatim}
\usepackage{graphicx}
\usepackage{booktabs}
\usepackage{amsmath}
\usepackage{bm}
\usepackage{cite}

\usepackage{threeparttable} % to use table notes
\usepackage{multirow}
\usepackage{color}

\def\BibTeX{{\rm B\kern-.05em{\sc i\kern-.025em b}\kern-.08em
    T\kern-.1667em\lower.7ex\hbox{E}\kern-.125emX}}
\usepackage{balance}

\usepackage[colorlinks=true, allcolors=blue]{hyperref}
\begin{document}

%\title{Learning Flight through Inclined Narrow Gaps with Onboard Sensing}
\title{Learning Agile Flights through Narrow Gaps with Varying Angles using Onboard Sensing}
% \title{Learning-based Agile Gap Traversal Flight Using Onboard Sensing}

\author{
    Yuhan Xie, Minghao Lu, Rui Peng and Peng Lu, Member, IEEE 
\thanks{
Manuscript created on December, 2022. 
This work was supported by General Research Fund under Grant 17204222, and in part by the Seed Funding for Strategic Interdisciplinary Research Scheme and Platform Technology Fund.
\textit{(Corresponding author: Peng Lu) }

The authors are with the Department of Mechanical Engineering, the University of Hong Kong, Hong Kong SAR, China 
(email: \{yuhanxie, minghao0, pengrui-rio\}@connect.hku.hk, lupeng@hku.hk).
}
\vspace{-0.8cm}
}

% \mark{IEEE/ASME TRANSACTIONS ON MECHATRONICS }% ~Vol.~18, No.~9, Septe mber~2020
% \markboth{IEEE/ASME TRANSACTIONS ON MECHATRONICS }% ~Vol.~18, No.~9, September~2020
% {How to Use the IEEEtran \ LaTeX \ Templates}

% \vspace{-0.4cm}
\maketitle

% \vspace{-0.4cm}
\begin{abstract}
% the problem
This paper addresses the problem of traversing through unknown, tilted, and narrow gaps for quadrotors using Deep Reinforcement Learning (DRL). Previous learning-based methods relied on accurate knowledge of the environment, including the gap's pose and size. In contrast, we integrate onboard sensing and detect the gap from a single onboard camera. The training problem is challenging for two reasons: a precise and robust whole-body planning and control policy is required for variable-tilted and narrow gaps, and an effective Sim2Real method is needed to successfully conduct real-world experiments. To this end, we propose a learning framework for agile gap traversal flight, which successfully trains the vehicle to traverse through the center of the gap at an approximate attitude to the gap with aggressive tilted angles. 
The policy trained only in a simulation environment can be transferred into different domains with fine-tuning while maintaining the success rate. Our proposed framework, which integrates onboard sensing and a neural network controller, achieves a success rate of 87.36$\%$ in real-world experiments, with gap orientations up to $60^\circ$. 
To the best of our knowledge, this is the first paper that performs the learning-based variable-tilted narrow gap traversal flight in the real world, without prior knowledge of the environment. 
%{87.36$\%$} in {87} traversals.
\end{abstract}

\vspace{-0.1cm}

%\section*{Supplementary Material}
%The accompanying video is available at:

\vspace{-0.4cm}
\section{Introduction}
% motivation
%% quadrotor, %% learning on robotics %% learning for quadrotor
Quadrotors are highly agile and versatile flying machines, making them ideal for complex tasks in cluttered environments \cite{lu2022perception}. 
Meanwhile, reinforcement learning (RL) is recently developing rapidly in the robotics domain for its strong potential of exploiting the robots' agility \cite{o2022neural, andrychowicz2020learning}. 
Therefore, research topics arise naturally to employ RL on quadrotors for aggressive tasks \cite{penicka2022learning, xiao2021flying}, which have recently contributed to a significant increase in autonomy capabilities \cite{loquercio2021learning}. 
%% learning SE(3) %% to narrow gap
Among the agile flight tasks in complex environments, one of the fundamental challenges is flying through narrow gaps, in which the drone's position and attitude must be considered simultaneously, leading to a \textit{SE(3)} planning and control problem. 
%the essential problem is the whole-body motion planning (\textit{SE(3)} planning) and control, which considers the position and attitude of the drone simultaneously, %to avoid complex obstacles. 

%% challenges of SE3 plan, motivation
% Some existing works consider learning to traverse narrow gaps for quadrotors. However, three main problems remain unsolved. 
Despite significant progress in learning-based gap traversing tasks, three main problems remain unsolved. 
%% 1. limited gap setting (narrow \& tilted); no good sim2real trials, lack of real-world exprs
% {1. variable angle}
%{2. onboard sensing}
%{3. imitation}
Firstly, training a policy in simulation and successfully transferring it to real-world flights through aggressive angle narrow gaps has not been addressed \cite{xiao2021flying, sun2022aggressive, chen2022learning}. 
%Firstly, transferring a policy trained in simulation to the real world for aggressive angle narrow gap traversal has not been addressed \cite{xiao2021flying, sun2022aggressive, chen2022learning}. 
The training algorithm is required to consider both an aggressive and robust \textit{SE(3)} control policy and an effective Sim2Real transfer. 
%% 3. unknown env. (onboard sensing)
Secondly, existing methods require prior knowledge of the gap pose and size in the world reference frame. 
% \textcolor{blue}{As these systems were not gap-aware, the trajectory or control commands cannot be re-generated for a variable angle gap.} 
Moreover, errors introduced by the gap detection would increase the risk of collision in real-world experiments. 
%% 2. with no expert experience, exploit the drl (not imitate)
Lastly, some approaches rely on expert planners and controllers for imitation in training \cite{lin2019flying}, which may end up with local optimal solutions similar to experts without sufficient exploration.
% \textcolor{magenta}{
% %A good initial condition could be provided by the 
% The prior expert knowledge could provide a good initial condition for the policy and accelerate the training process. 
% However, imitation learning may end up with local optimal solutions that are similar to priors without sufficient exploration. %and limits the exploration ability of RL. 
% }

% limited gap setting (narrow \& tilted); no good sim2real trials, lack of real-world exprs
% with no expert experience, exploit the drl (not imitate)
% unknown env. (onboard sensing)

\begin{figure}[t] 
\centering
\includegraphics[width=\linewidth]{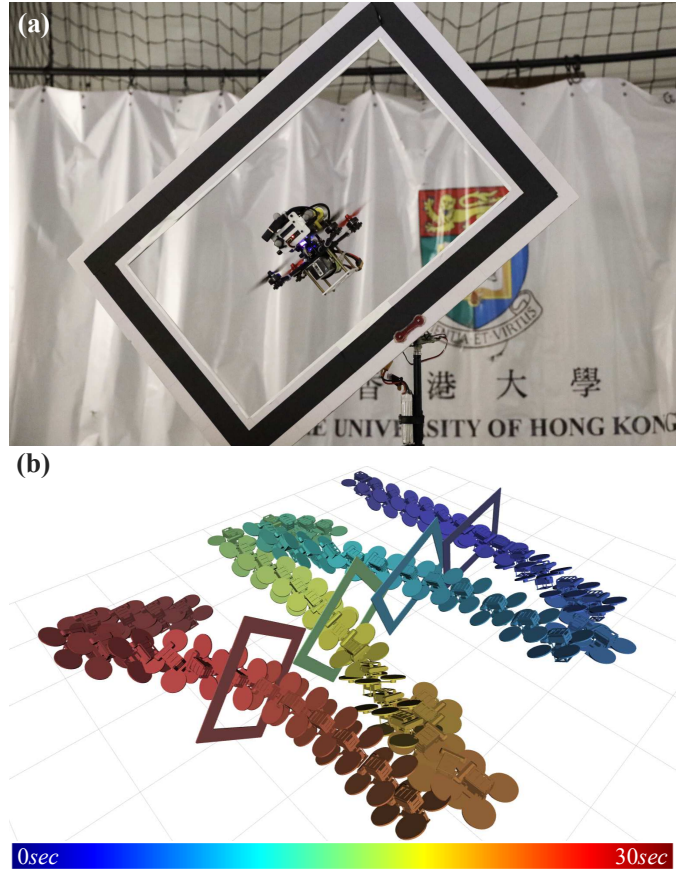}
\caption{
Agile flights through tilted narrow gaps in real-world experiments. 
(a) Our quadrotor is traversing a tilted narrow gap. 
(b) Experiment results of four consecutive flights through narrow gaps. Each traversal is translated as a whole for visualization. 
}
\label{fig_traversing}
\vspace{-0.5cm}
\end{figure}

To overcome the aforementioned challenges, this paper proposes an end-to-end framework that includes a gap detection algorithm and a policy training algorithm, 
%enabling autonomous gap detection and traversal. 
%which autonomously detects the pose of the gap to traverse and control the vehicle fly through. 
%In our work, the gap to traverse is narrow and has a variable tilted attitude. 
which enables quadrotors to autonomously detect and traverse gaps with variable-tilted attitude. 
The training algorithm takes generalization and domain adaption into account, thereby ensuring successful Sim2Real transfer for physical experiments. 
%The method requires no prior information about the narrow gap, making it applicable to unknown environments. 
%Requiring no prior information (e.g., position, orientation, size) of the narrow gap, our method is applicable to unknown environments of a window on a wall. 
% 4. summary
The main contributions of our work are summarized as below:
\begin{enumerate}
    \item % design a training method
    A novel learning framework is designed for variable-tilted narrow gap traversing tasks. 
    The trained policy achieves a precise \textit{SE(3)} trajectory planning and control of a quadrotor. 
    %A multi-stage curriculum learning method, a concise reward function and an augment on policy input are designed to fulfill the delicate task. 
    \item % stability of the policy
    With fine-tuning to transfer the policy from the training environment, repetitive tests in the software-in-the-loop (SITL) environments are conducted, maintaining a high success rate and demonstrating the effectiveness of the training algorithm. 
    %With a little fine-tuning to transfer the policy from the training environment, repetitive tests in both simulation and real world are conducted, demonstrating the effectiveness of the trained policy. 
    %The high successful rate, 72.69\% in the software-in-the-loop (SITL) experiment, shows the stability of our training algorithm. 
    \item % propose/ build a system: learning + Onboard sensing ... 
    Onboard sensing is introduced so that no prior knowledge of the gap is required, e.g., position, orientation, or size. 
    % We introduce onboard sensing technique with the neural network controller so that no prior knowledge of the gap is required (e.g. position, orientation, size). 
    % The system uses depth camera to estimate the gap position and orientation, so that no prior knowledge of the gap is required. 
    % Our system integrates perception, planning and control as an end-to-end pipeline, which can exploit the gap traversing task autonomously. %independently.
    To the best of our knowledge, this is the first work that integrates onboard sensing to a learning system for gap traversing tasks. 
    \item %experiment
    Repetitive real-world experiments demonstrate the robustness of the proposed framework. 
    Our experiment results show that our quadrotor system can fly through variable-tilted narrow gaps with precise traversing posture for gap orientations up to $60^\circ$. % (average position error $<9$cm, average attitude error $<12^\circ$) in over 70 traversal flights. 
\end{enumerate}
% \end{itemize}

%The rest of this letter is organized as follows. Section II introduces the related works...

\vspace{-0.2cm}
\section{Related Work}\label{sec_related_work}
% Traditional method (agile flight, gap traversing)
\subsection{Quadrotor Agile Flight}
State-of-the-art agile quadrotor flight methods typically decouple trajectory planning and control. 
For a specific environment, the conventional approach usually follows the pipeline of planning a trajectory and then tracking the trajectory by the controller. 
The performance and success rate depend highly on both the quality of the planned trajectory and the robustness of the controller. 
For quadrotor trajectory generation, the modern frameworks exploit the differential flatness \cite{mellinger2011minimum} of the vehicle using polynomial \cite{mueller2015computationally, ren2022bubble, richter2016polynomial}, or B-spline \cite{penin2018vision, zhou2021raptor, usenko2017real} representations.  
These trajectories are inherently smooth. 
Hence, they cannot represent the rapid state or input changes in a reasonable order, and only reach the input limits for an infinitesimal short duration \cite{penicka2022learning}. 
The popular controllers for trajectory tracking include model predictive control (MPC) and differential flatness control \cite{mueller2015computationally, ren2022bubble, richter2016polynomial, penin2018vision}. 
However, most control approaches rely on physical assumptions and are dependent on modeling, making them  struggle to handle disturbances during agile flight. 

% learning-based agile flight
In contrast to the decoupled framework in optimization-based methods, learning-based methods address the problem by learning an end-to-end policy that predicts control commands directly from high-dimensional observations \cite{penicka2022learning, hwangbo2017control, chen2022learning, lin2019flying, xiao2021flying}. 
Recent works have shown that these methods can achieve superhuman performance in near-time-optimal flight for drone racing and high-speed flight in the wild \cite{loquercio2021learning}. 

\vspace{-0.2cm}
\subsection{Agile Flight through Narrow Gaps}
% traditional methods for gap traversing problem ....
Aggressive flight through a narrow gap is one of the most challenging problems for quadrotors. 
A whole body planning and control considering position as well as attitude of the vehicle is required. 
Early work designed a sequence of control phases to execute an aggressive trajectory and reach the goal state \cite{mellinger2012trajectory}. 
% Early work decouples the control framework into position and attitude control, and designs a sequence of trajectories offline to reach the goal state \cite{mellinger2012trajectory}. 
Based on the differential flatness property \cite{mellinger2011minimum}, Loianno \textit{et al.} \cite{loianno2016estimation} planned dynamically feasible trajectories which guide the drone to the window traversal state. 
The work also considers state estimation from a monocular camera and an IMU. 
Falangal \textit{et al.} \cite{falanga2017aggressive} further integrated state estimation and gap detection by onboard sensing and computing, and achieved the goal without prior knowledge of the pose of the gap. 

% learning-based method
Recently, some works have considered learning-based planning and control methods to address the gap traversal problem for quadrotors. 
Early work \cite{lin2019flying} follows the decoupled planning and control pipeline, and imitates a traditional planner \cite{mueller2015computationally} and controller \cite{mellinger2011minimum,lee2010geometric}. 
Additional reinforcement training is also required to fine-tune the policy network. 
%This work indicates that learning-based approaches can address the problem of flying through tilted narrow gaps and has the potential to achieve higher performance in some aspects.
%% %%However, providing prior knowledge of plan and control is not only time-consuming but also limits the exploration ability of RL. 
%However, imitation learning may end up with local optimal solutions and limits the exploration ability of RL. 
The prior expert knowledge provides good initial conditions for the policy and accelerates the training process. 
However, the imitation learning may end up with local minimums similar to priors, which limits the exploration ability of RL. 
Moreover, the control command of desired attitude generated by the policy vibrates severely compared to the result of the traditional, indicating an unsatisfying control performance.  
To exploit the quadrotors' agility, recent work employs deep reinforcement learning for the gap traversal problem \cite{xiao2021flying, chen2022learning, sun2022aggressive}. 
Our previous work \cite{xiao2021flying} proposed a reinforcement learning framework augmented with curriculum learning and Sim2Real methods, which achieves successful real-world gap traversing flight using DRL. 
However, the tilted angle of the gap is fixed at $20^\circ$ in training and experiments. 
% The tilted angle is also limited in \cite{sun2022aggressive}. 
Chen \textit{et al.} \cite{chen2022learning} considered narrow gaps with up to $60^\circ$ tilted angle in simulation, while the physical experiments were conducted with a very limited tilted angle. 
A successful Sim2Real transfer is not presented for aggressive angles. 
Overall, a learning-based control policy for traversal through aggressive angle gaps in the real world remains unsolved among these works. 
To tackle this problem, our training algorithm considers not only the aggressive and robust \textit{SE(3)} control but also the effective Sim2Real transfer. 
% gap detection
Furthermore, the related work mentioned above relied on accurate prior knowledge of the gap, including the position, orientation, and size. 
Thus, these methods cannot address the problem when the gap state changes. 
In this work, we introduce an onboard sensing algorithm to detect the gap, which is necessary for real-world applications. % in unknown environments. 
%The onboard sensing and a gap-aware system is necessary for the real-world application in unknown environments. 

%[Figure.4. overall architecture]
% \begin{figure*}[t] 
% \begin{figure}[t] 
% \centering
% \includegraphics[width=\linewidth]{figure_method/overall_archi_v5_2.pdf}
% \caption{The overall architecture in physical experiments.}
% \label{fig_overall_archi}

% \vspace{-0.6cm}
% \end{figure}

\vspace{-0.2cm}
\section{Problem Statement}\label{sec_problem} 
% Problem overview % difficulty to be solved % state the problem physically
In this letter, we address the problem of controlling a quadrotor to fly through a narrow gap with varying tilted angles using an onboard camera. 

\vspace{-0.2cm}
\subsection{Problem Overview}
\begin{figure}[t] 
\centering
\includegraphics[width=0.7\linewidth]{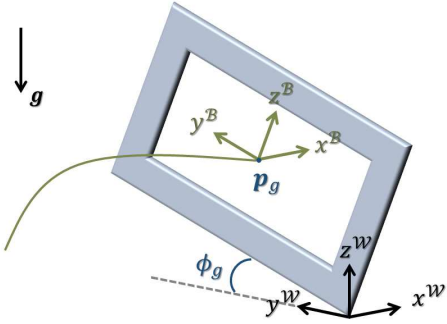}
\caption{Traversal Demonstration. }
\label{fig_traverse_demo}
\vspace{-0.4cm}
\end{figure}

Our approach consists of two subsystems: perception and control. 
The perception system estimates the position and orientation of the gap using a forward-facing depth camera, which is presented in Section \ref{sec_sensing}. 
The control system includes a neural network that maps from the observation of the drone and gap, directly to low-level control commands, guiding the quadrotor to complete the task. 
The trajectory should try to intersect the center of the gap while simultaneously attaining the exact orientation of the gap, as illustrated in Figure \ref{fig_traverse_demo}. 
Therefore, a precise \textit{SE(3)} planning and control policy for quadrotor is required. 

Variation of gap orientation is considered. 
In policy training, we keep the drone facing the gap and omit the yaw angle control. 
Pitch angles of the gap are ignored, as the gap on a wall usually has a few pitches. 
Thus, we mainly cope with the variation of roll angle in this letter.

\vspace{-0.2cm}
\subsection{Quadrotor Dynamics for Training}
To simulate the quadrotor flight and the interaction between the vehicle and the gap for policy training, we formulate the quadrotor model. 
%This section introduces the quadrotor model that will be used for the training. 
Consider a quadrotor with mass $m\in \mathbb{R}$ and diagonal moment of inertia matrix $\bm J \in \mathbb{R}^{3\time3}$. 
% Define an inertial reference frame denoted by $[\bm e_1,\bm e_2,\bm e_3]$ and a body reference frame centered in the center of mass of the quadrotor denoted by $\bm R=[\bm b_1, \bm b_2, \bm b_3] \in SO(3)$.
The dynamic model of the system can be written as
\vspace{-0.1cm}
\begin{equation}
\begin{aligned}    \label{eq_drone_model} 
    \dot{\bm p} &=\bm v,  	&\,
    m\dot{\bm v} &= \bm R\bm e_3 f_T + \bm R\bm f_D +m\bm g 	 \\
    \dot{\bm R} &= \bm R\hat{\bm\omega}, 	&\,
    \bm J\dot{\bm\omega} &= -{\bm\omega}\times\bm J\bm\omega + \bm\tau_T + \bm\tau_D  
\end{aligned}
\vspace{-0.1cm}
\end{equation}
where $\bm p =[p_x, p_y, p_z]^T$ and $\bm v$ are the position and velocity vector in the world frame, 
$\bm R \in \mathbb{SO}(3)$ is the rotation of the quadrotor, $\bm\omega$ represents angular body velocity. 
$\hat{\bm\omega}$ is the skew-symmetric matrix of vector $\bm\omega$, $\bm g$ is the gravity vector, and $\bm e_3 =[0, 0, 1]^T$ is a constant vector. 
$f_T$ and $\bm\tau_T$ denote thrust in the body-z axis and body torque generated by four rotors. 
Air drag force $\bm f_D$ and torque $\bm\tau_D$ are also modeled for aggressive motion. 
Overall, the state and control input of quadrotor are $\bm x = [\bm p, \bm v, \bm R, \bm \omega]^T$, $\bm u = [f_T, \bm\tau_T]^T$. 
We define the Euler angles of the quadrotor $(\phi, \theta, \psi)$, which can be derived from $\bm R$.

\vspace{-0.2cm}
\subsection{Task Formulation}\label{sec_sub_task_formu}
% formulate the problem in DRL framework.
%We formulate the tilted narrow gap traversing control problem into a deep reinforcement learning framework. 
%Assume that a quadrotor flying in a workspace $\mathcal{S}$ with a tilted narrow window on a wall. 
%The state of the gap $\bm g\in\mathcal{G}$ can be observed by an onboard depth camera

We model the task using an infinite-horizon Markov Decision Process (MDP), defined by the tuple $(\mathcal{S}, \mathcal{A}, p, r)$, where the state space $\mathcal{S}$ and the action space $\mathcal{A}$ are continuous. 
At every control step $t$, given current state $\bm s_t\in\mathcal{S}$, an action $\bm a_t \in \mathcal{A}$ is sampled from a policy $\pi(\bm a_t|\bm s_t)$. 
Subsequently, the agent executes the action $\bm a_t$ and transits to the next state $\bm s_{t+1}\in\mathcal{S}$ with the unknown state transition probability $p:\mathcal{S}\times\mathcal{S}\times\mathcal{A}\rightarrow\left[ 0,\infty\right) $, receiving a bounded reward $r:\mathcal{S}\times\mathcal{A}\rightarrow[r_{\rm{min}},r_{\rm{max}}]$. 
Specifically, the state $\bm s\in\mathcal{S}$ includes the quadrotor state $\bm x\in\mathcal{X}$ and the gap pose $\bm g\in\mathcal{G}$. The goal of our algorithm is to learn a control model $\pi : \mathcal{X}\times\mathcal{G}\rightarrow\mathcal{A}$.

%Specifically, the state $\bm s\in\mathcal{S}$ includes the quadrotor state and gap information. 
%Let $\mathcal{X}$ be the space of quadrotor state, and $\mathcal{G}$ be the space of gap pose (including position and orientation). The goal of our algorithm is to learn a control model $\pi : \mathcal{X}\times\mathcal{G}\rightarrow\mathcal{A}$. 
%The model $\pi$ takes a full-state of quadrotor $\bm x\in\mathcal{X}$ and a gap pose $\bm g\in\mathcal{G}$ as input, and generates a low-level action $\bm a\in\mathcal{A}$ to the drone.

\vspace{-0.2cm}
\section{Learning to Control}\label{sec_learning}
%Policy Training Methodology
This section presents the policy architecture, reward formulation, and training strategy employed in our approach for training a control policy for the tilted narrow gap traversal problem. 

\vspace{-0.2cm}
\subsection{Policy Architecture}
The neural network architecture as well as the state and action spaces are illustrated in Figure \ref{fig_network_arch}. 
%demonstrates , as well as , which is a 2-layer multi-layer perception (MLP). 
An additional $\tanh$ function is used at the last layer of the policy network to keep the actions within a fixed range. 

\subsubsection{States}
As stated in Section \ref{sec_sub_task_formu}, the state space of our neural network consists of two parts: the quadrotor state and the gap pose. 
We define the gap pose by center position $\bm p_g\in\mathbb{R}^3$ and the rotation matrix $\bm R_g\in\mathbb{SO}(3)$ in the world frame. 
The corresponding Euler angles are $(\phi_g, \theta_g, \psi_g)$.
%Furthermore, we denote $(\phi_g, \theta_g, \psi_g)$ as the Euler angles of the gap related to $\bm R_g$. 
%Specifically, let $\bm p_g\in\mathbb{R}^3$, $\bm R_g\in\mathbb{SO}^3$, and $(\phi_g, \theta_g, \psi_g)$ be the center position, rotation matrix and Euler angles of the gap. 
%$\pmb{g} = [p_{g,x}, p_{g,y}, p_{g,z}, \phi_g, \theta_g, \psi_g]^T $. 

%\subparagraph{Drone State} 
\textit{Drone States.} 
To facilitate traversal, the pose information of the quadrotor is given relative to the target. 
We denote $\bm p_T$ as a target position located behind the gap center that
\vspace{-0.1cm}
\begin{equation}\label{eq_target_position}
	\bm p_T = \bm p_g + \delta_T \cdot\bm R_g \bm e_1
\vspace{-0.1cm}
\end{equation}
where $\delta_T$ is a target distance to the gap center and $\bm R_g \bm e_1$ represents the first column of the $\bm R_g $.
The relative position vector $\bm p^e$ is designed as
\vspace{-0.1cm}
\begin{equation}
	p^e_i = {\rm sgn}( p_{T,i} - p_i)\sqrt{| p_{T,i} - p_i|},\quad i \in\{x,y,z\}
\vspace{-0.1cm}
\end{equation}
The relative orientation is defined by subtracting the Euler angles of gap and quadrotor as
\vspace{-0.1cm}
\begin{equation}
	\phi^e = \phi_g - \phi,\quad \theta^e = \theta_g - \theta
\vspace{-0.1cm}
\end{equation}
Although the subtraction is physically meaningless, it is intuitive for policy training, guiding the quadrotor approaches gap's roll and pitch angle during traversal.

\begin{figure}[t] 
\centering
\includegraphics[width=\linewidth]{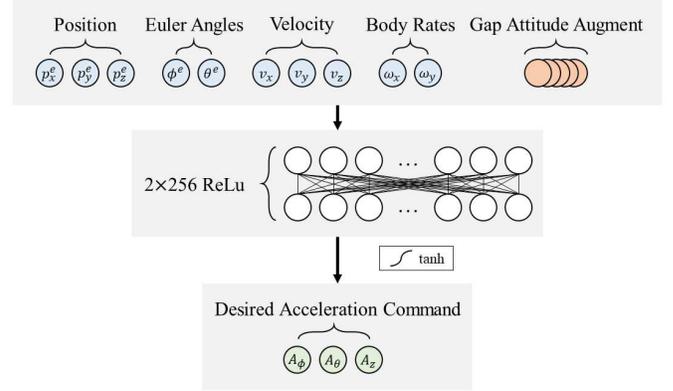}
\caption{Neural Network Architecture.}
\label{fig_network_arch}

\vspace{-0.4cm}
\end{figure}

% augmentation of roll angle
\textit{Gap Attitude Augment. } 
Previous works only considered limited tilted angles in policy training or experiments. 
In contrast, we are interested in the variation of the gap orientation. %, especially for roll angle.
Therefore, we implement a data augmentation technique on the state-based inputs to improve the data efficiency as well as the generalization ability of the policy \cite{laskin2020reinforcement}. 
Specifically, the random amplitude scaling method is introduced in this work for gap attitude, as shown in Figure \ref{fig_network_arch}.

%% Figure. policy network and value network Structure

\subsubsection{Actions} 
Network actions are normalized second-order derivatives of desired Euler angles and altitude, while the low-level control commands for the vehicle are the desired orientation and altitude. 
Hence, after mapping the normalized network outputs to a fixed range, there is a second-order integrator before passing the signals to the low-level controller on the quadrotor. %, as illustrated in Figure \ref{fig_overall_archi}. 

There are two considerations for this design of network outputs. 
% smooth the network outputs
Firstly, the network outputs are physically meaningful and effective for agile quadrotor flight control. 
% Based on the differential flatness of quadrotor dynamics \cite{mellinger2011minimum}, the second derivatives of orientation appear as functions of the fourth derivatives of the position, i.e., the snap, and also related to the control inputs $\tau_{T,x}$, $\tau_{T,y}$, $\tau_{T,z}$, as defined in \eqref{eq_control_input}. 
Based on the differential flatness of quadrotor dynamics, the control inputs $\bm\tau_{T}$ appear as functions of the second derivatives of orientation, 
%Also, the second derivatives of height appear as functions of the control input $f_T$. 
and $\bm f_T$ appears as the function of the second derivatives of altitude. 
Thus, our policy can be considered as a motion planner on thrust and torque, which has been demonstrated effective for agile quadrotor flight planning. 
%Meanwhile, the integrator can further smooth the network outputs, giving a better control performance on the vehicle. 
Secondly, the network output processing framework can facilitate Sim2Real transfer, referring to our previous work \cite{xiao2021flying} which demonstrated the framework could enhance generalization without utilizing real-world data.

\vspace{-0.2cm}
\subsection{Reward Function Design}
% \vspace{-0.1cm}
The reward function consists of four designs. 
The main objective of our task is to guide the drone to fly to the back side of the gap. 
The position distance between the quadrotor position $\bm p$ and the target point $\bm p_T$ defined by \eqref{eq_target_position}, is calculated as the position reward as follows
\vspace{-0.1cm}
\begin{equation}
\label{eq_r_p}
	r_p(t) = -\left\| \bm p(t) - \bm p_T(t)\right\| 
\vspace{-0.1cm}
\end{equation}
Meanwhile, to increase the margin between gap while traversal, the quadrotor should raise its roll angle to the same attitude as the gap and reduce its pitch angle to zero, as illustrated in Figure \ref{fig_traverse_demo}. 
Thus, we design an attitude reward of relative roll between quadrotor and gap when the vehicle approaches the gap. 
\vspace{-0.1cm}
\begin{equation}
\label{eq_r_a}
r_a(t) = 
\begin{cases}
    - \min( \tan \left|\phi^e \right|, 50)  & \text{approach gap} \\
    0 & \text{otherwise}
\end{cases}
\vspace{-0.1cm}
\end{equation}
Note that there is no constraint on pitch, leaving more space for policy exploration. 
A penalty on control input is also given for smooth control
\vspace{-0.1cm}
\begin{equation}
\label{eq_r_u}
	r_u(t) = -\left\| \bm u(t) \right\| 
\vspace{-0.1cm}
\end{equation}
Lastly, a terminal reward $r_T$ is given only when the vehicle successfully passes through the window without any collision detected. 
The total reward $r(t)$ at time $t$ is defined as
\vspace{-0.1cm}
\begin{equation}\label{eq_reward_total}
r(t)\!=\!\lambda_p r_p(t) +\!\lambda_a(r_a(t)\!+\!b_a) +\!\lambda_u r_u(t) +\!
\begin{cases}
    r_T &\! \text{win}       \\
    0   &\! \text{otherwise}
\end{cases}
\vspace{-0.1cm}
\end{equation}
where $\lambda_p, \lambda_a, \lambda_u \in\mathbb{R}$ are hyperparameters that trade-off between each reward components, $b_a\in\left[0,+\infty\right) $ is a positive offset for relative attitude reward.

\vspace{-0.2cm}
\subsection{Training Details} 
\label{sec_sub_train_detail}   
% algorithm
The policy is trained using Soft Actor-Critic (SAC) \cite{haarnoja2018soft}, an off-policy algorithm that features entropy regularization. 
% how we train
% In our training environment, a quadrotor with dynamics \eqref{eq_drone_model}\eqref{eq_rotor2thrust&torque} and a static window on a wall are simulated. 
In our training environment, a quadrotor with dynamics \eqref{eq_drone_model} and a static window on a wall are simulated. 
The vehicle is simulated at a frequency of $80\rm{Hz}$, while the control frequency, i.e., the frequency of collecting state and action data for training, is only $20{\rm Hz}$, which balances the training acceleration and data efficiency. 
The episodes terminate when the edge of state space $\mathcal{S}$ is reached, or the terminal reward $r_T$ is obtained.  

\subsubsection{Curriculum Learning}
The terminal reward is hard to obtain directly due to the narrow gap. 
Only a precise control policy can complete the task and win the terminal reward. % while winning the terminal reward. 
To overcome reward sparsity, a curriculum strategy is employed for policy training in multiple stages.
Specifically, we refer to our previous work \cite{xiao2021flying} and introduce a difficulty factor $d_f$ to adjust gap size with training episodes. 
As training episodes increase, the gap narrows so the feasible traversal trajectories converge. 
% As episode increases, gap size decreases, and the traversal state for quadrotor converges. 
%The roll angle of the gap $\phi_g$ is randomly given in uniform distribution initially. 
To augment the policy for aggressive cases ($|\phi_g|\geq50^\circ$), after the gap shrinks to the goal size, we further add a curriculum that makes the probability of large roll angles greater. 
%We changed the distribution of the roll angle, so that the probability of large angles is greater

\subsubsection{Randomization}
Several randomization strategies are employed to make the policy robust against unknown dynamics effects and facilitate domain adaptation. 
% how we simulate the drone and window % generalization
For each episode, the vehicle and the window are initialized with randomization: the initial state of quadrotor is normally distributed. 
%, and the roll angle of the gap $\phi_g$ is randomly given in uniform distribution. 
The dynamics parameters of vehicle are also randomized with normal distributions. %(model inaccuracy). 
%The expectations of dynamics parameters are given in Table \ref{tab_drone_dynamic_params}. 
For each step, observation noises are introduced in zero-mean normal distributions. %(uncertainty of sensors)
%The expectations or standard variations of the distributions can be found in [Table 1].

\vspace{-0.2cm}
\section{Onboard Sensing}\label{sec_sensing}
% \vspace{-0.1cm}
% In this section, we present our method of gap detection and state estimation with a RGBD sensor. In the sim-to-real process of our planning and control policy, the narrow is implemented as a random size black-and-white rectangular box frame. The overall description of the problem is shown in Figure \ref{fig_traverse_demo}. %\ref{fig_sensing}.
%This section presents the gap detection method. % by an RGBD sensor. 
%In the physical experiments, the gap is configured as a black-and-white rectangular frame with an uncertain size. An RGB image and a depth image can be obtained from an RGB-D camera. The overall gap detection and pose estimation framework is shown in Figure \ref{fig_overall_archi}. 

This section introduces the gap detection method, which aims to identify the black-and-white rectangular frames with uncertain sizes in physical experiments. 
The method employs an RGB-D camera to obtain both an RGB image and a depth image. 
% The overall framework for gap detection and pose estimation is illustrated in Figure \ref{fig_overall_archi}. 

%Firstly, we run the Canny detector on the binary image after close operation, undistort all edges, and group close edges. Among the edge groups, we fit them to polygons based on Douglas Peucker algorithm \cite{ramer1972iterative}, and then generate the convex hulls of the polygons by Quickhull algorithm\cite{barber1996quickhull}. 
%Among the convex hulls, a rectangle can be determined by the following conditions

% get edge, get convex hull
% We first identify rectangles.  
To extract and refine the edges from the binary image, we perform the closing operation, Canny edge detection, edge undistortion, and edge grouping consecutively. 
Subsequently, we apply the Douglas-Peucker algorithm to fit the edge groups into polygons, followed by generating their respective convex hulls using the Quickhull algorithm. 
The convex hull $\mathcal H_k$ is defined as a set of pixels $[u, v]^T$, given by
\vspace{-0.1cm}
\begin{equation}
\mathcal H_k= \left\{[u_n, v_n]^T\mid n=1,2,\dots,N \right\}
\vspace{-0.1cm}
\end{equation}
%The convex hull $\mathcal H_k$ is defined as a set of pixels $(u, v)$. %%%%
% get rectangles
To identify rectangles among the convex hulls, we consider the following conditions: 
\vspace{-0.1cm}
\begin{equation}
\begin{split}
        & N = 4, \  \left({\rm arccos}(\bm e \cdot \bm e_a) - 1\right)^2 < \epsilon_1
\end{split} 
\vspace{-0.1cm}
\end{equation}
where $\bm e$ and $\bm e_a$ denote adjacent edge vectors of a hull, and $\epsilon_1$ is a small constant factor. 
% the number of vertices is four, and the angle between adjacent edges is smaller than a threshold $\epsilon_1$.  %%%%

% 3D position of rectangles
% After that we can get depth of the four vertices of each rectangle by the aligned depth image. Here we can obtain the pixel position (u, v) and depth d of each vertex
We proceed by estimating the 3D positions of the detected rectangles using the aligned depth image. 
The depth $d$ of each vertex within a rectangle is obtained by providing its pixel position $[u,v]^T$. 
Given the camera intrinsic matrix $\bm M_{1}\in\mathbb{R}^{3\times 4}$ and the world-to-camera transformation matrix $\bm M_{2}\in\mathbb{R}^{4\times 4}$, the 3D position of each vertex $\bm p_v=[x_v,y_v,z_v]^T$ in the world frame can be calculated by
\vspace{-0.1cm}
\begin{equation}
\label{eq_cam_matrix}
\bm M_1 \bm M_2 [x_v,y_v,z_v,1]^T = d[u,v,1] ^T
\vspace{-0.1cm}
\end{equation}
%Thus, each detected rectangle can be formulated as a point set of its four vertices as
%As a result, we can represent each detected rectangle as a set of four vertices in 3D space:
Hence, each detected rectangle can be represented as a set of four vertices in 3D space:
\vspace{-0.1cm}
\begin{equation}
\label{eq_p_set}
%    \xi = \left\{ \bm p_{v,j}\mid j\in(1,4) \right\}
    \xi = \left\{ \bm p_{v,j}\mid j =1,2,3,4 \right\}
\vspace{-0.1cm}
\end{equation}
% gap identify
%Then, the outlines of the gap can be determined by finding the two rectangles that hold the following conditions,
%To determine the outlines of the gap, we find the two rectangles that satisfy the conditions:
The outlines of the gap can be determined by finding two rectangles $\xi_1$, $\xi_2$ that satisfy the following conditions,
\vspace{-0.1cm}
\begin{equation}
\label{eq_conditions}
\begin{split}
	{\rm arccos}(\bm{n}_1 \cdot \bm{n}_2)  < \epsilon, \\
	{\rm area}(\xi_1) \subseteq {\rm area}(\xi_2)
\end{split} 
\vspace{-0.1cm}
\end{equation}
where $\bm n_i$ denotes the normal unit vector of a rectangle plane, and $\epsilon$ is a small constant factor. 
The term $\rm area(\xi)$ refers to the area confirmed by the points in $\xi$. 
These conditions, as stated in \eqref{eq_conditions}, describe the relationship between the inner $\xi_2$ and outer $\xi_1$ rectangular borders of the gap, which should be in the same plane, and the inner area is a proper subset of the outer area.
% The conditions \eqref{eq_conditions} describe the relationship between the inner and outer rectangular borders of the gap, specifically in the same plane and the inner area is the proper subset of the outer area.

% gap pose

%Therefore, from the two point sets $\xi_{1}$, $\xi_{2}$ defined by \eqref{eq_p_set}, the 6-DOF pose of the gap can be obtained by geometry calculation. 
%\begin{equation}
%\bm p_g = \frac{1}{8} \sum \limits_{i=1}^{8} \bm p_{i}, \quad \bm p_{i} \in \xi_{1} 
%\cup 
%\xi_{2}
%\end{equation}
%and the attitude of the gap can be represented as a rotation matrix
%\begin{equation}
%\bm R_g = \left[\bm{r}_{x}, \bm{r}_{y}, \bm{r}_{z} \right]
%\end{equation}
%where $\bm{r}_{x}$ and $\bm{r}_{y}$ are the unit vectors towards the direction of the long side and the short side of the rectangle in world frame, $\bm{r}_{z}$ is the normal unit vector of the gap plane in world frame. 

The 6-DOF pose of the gap can be calculated from the two point sets $\xi_{1}$ and $\xi_{2}$ using geometry calculations. 
The central position of the gap, $\bm p_g$, is determined as the average of the vertices, 
while the rotation matrix of the gap, $\bm R_g$, is calculated using the sides of the rectangle and the normal vector of the rectangle plane. 
The rotation matrix $\bm R_g$ is then transformed to Euler angles $(\phi_g, \theta_g, \psi_g)$.
%After that, we can transform the rotation matrix $\bm R_g$ to Euler angles $(\phi_g, \theta_g, \psi_g)$.
%\begin{equation}
%\label{eq_filter}
%\left[\hat{\bm{p_{g}}}, \hat{\theta}, \hat{\psi}, \hat{\phi} \right] = 
%\sum \limits_{m=1}^{M} \omega_{m} \left[\bm p_{g, m}, \theta_{g,m}, \psi_{g,m}, \phi_{g,m} \right], %\sum \limits_{m=1}^{M} \omega_{m} = 1
%\end{equation}
%where M is the window size. 
Overall, we use $\bm x_g = [\bm p_g, \phi_g, \theta_g, \psi_g]^T$ to describe gap pose. 
To smooth the output gap pose, a third-order low-pass filter is applied using the following equation,
\vspace{-0.1cm}
\begin{equation}
\begin{split}
    \bm {\dot x}_1 =& \bm x_2 \\
    \bm {\dot x}_2 =& \bm x_3 \\
    \bm {\dot x}_3 =& \omega_1\omega_2^2(\bm x_{g,m} - \bm x_1) \\
    &- (2\zeta\omega_1\omega_2+\omega_2^2)\bm x_2 - (\omega_1+2\zeta\omega_2)\bm x_3
\end{split}
\vspace{-0.15cm}
\end{equation}
where $\bm x_1 = \bm{\hat x}_g, \bm x_2 = \hat{\dot{\bm{x}}}_g, \bm x_3 = \hat{\ddot{\bm{x}}}_g$. 
The transfer function of this filter is $\omega_1\omega_2^2/(s+\omega_1)(s^2+2\zeta\omega_2s+\omega_2^2)$. 
The gap detector operates at a frequency of 30 Hz in our system.
%The gap detector runs at 30 Hz in our system. 

%% Training Parameters
\begin{table}[]%[t]			% ok
\centering
\caption{
Parameters of Training Algorithm}
\label{tab_train_params}
\begin{tabular}{l|ll}
\toprule
    &\textbf{Parameter}            					& \textbf{Value}		\\ 
\midrule
    \multicolumn{1}{l|}{\multirow{8}{*}{ \rotatebox{90}{RL} }}    
                                & position reward coefficient ($\lambda_p$)             &	$1.0$     \\
    \multicolumn{1}{c|}{}       & attitude reward coefficient ($\lambda_a$)             &	$10.0$				\\
    \multicolumn{1}{c|}{}       & attitude reward offset ($b_a$)						&	$0.2$				\\
    \multicolumn{1}{c|}{}       & control input reward coefficients ($\lambda_u$)		&	$0.05$				\\
    \multicolumn{1}{c|}{}       & terminal reward ($r_T$)								&	$500$				\\
    \multicolumn{1}{c|}{}       & window roll range 									& 	$[-60^\circ,+60^\circ]$	\\
    \multicolumn{1}{c|}{}       & target distance ($\delta_T\,{\rm m}$)			        &	$0.25$		\\
    \multicolumn{1}{c|}{}       & network outputs mapping scale ($\bm\kappa$)           &	$[80, 80, 24] $		\\
    %drone size				&	$0.33{\rm m} \times0.33{\rm m} \times 0.17{\rm m}$	\\
    %initial gap size		&	$1.50{\rm m} \times 0.90{\rm m}$					\\
    %final gap size			&	$0.50{\rm m} \times 0.25{\rm m}$					\\
\midrule
    \multicolumn{1}{l|}{\multirow{7}{*}{ \rotatebox{90}{SAC\cite{haarnoja2018soft}} }}    
                                & optimizer                & Adam \\%\cite{kingma2014adam} \\
    \multicolumn{1}{c|}{}       & learning rate                       	    & $3\times10^{-4}$     \\
    \multicolumn{1}{c|}{}       & discount factor ($\gamma$)			    & 0.95                 \\
    \multicolumn{1}{c|}{}       & replay buffer size                        & $10^{5}$             \\
    \multicolumn{1}{c|}{}       & batch size                                & 512                  \\
    \multicolumn{1}{c|}{}       & target smoothing coefficient ($\tau$)     & 0.01                 \\
    \multicolumn{1}{c|}{}       & target update interval              	    & 16                   \\ 
\midrule
    % \multicolumn{1}{l|}{\multirow{8}{*}{ \rotatebox{90}{Quadrotor} }}        
    %                         & $m\,[\rm kg]$                       	& $1.1$					\\%mass 
    % \multicolumn{1}{c|}{}   & ${\rm diag}(J)\,[\rm kg\,m^2]$ 		& $[0.12, 0.12, 0.22] $	\\%moment of inertia 
    % \multicolumn{1}{c|}{}   & $[\omega_{r,\rm{min}},\omega_{r,\rm{max}}]$		& $[50, 2000]$     		\\%rotor speed limit 
    % \multicolumn{1}{c|}{}   & $k_T$									& $6\times10^{-6}$      \\%thrust coefficient 
    % \multicolumn{1}{c|}{}   & $k_{TQ}$								& $0.02$      			\\ % moment to thrust coefficient 
    % \multicolumn{1}{c|}{}   & $l\,[\rm m]$							& $0.34$     			\\%arm length 
    % \multicolumn{1}{c|}{}   & $\bm k_{f_D}\,[\rm N\,s^2\,m^{-2}]$    & $[2.9, 2.9, 5.7]\times10^{-2}$    \\%air drag force coefficient 
    % \multicolumn{1}{c|}{}   & $\bm k_{\tau_D}\,[\rm N\,s^2]$	    & $[3.2, 3.2, 1.7]\times10^{-3}$    \\%air drag torque coefficient 
    \multicolumn{1}{l|}{\multirow{5}{*}{ \rotatebox{90}{Quadrotor} }}        
                            & mass ($m\,[\rm kg]$)                      	            & $1.1$					       \\ 
    \multicolumn{1}{c|}{}   & moment of inertia (${\rm diag}(J)\,[\rm kg\,m^2]$)        & $[0.12, 0.12, 0.22] $	    \\
    \multicolumn{1}{c|}{}   & thrust coefficient ($k_T$)                                & $6\times10^{-6}$          \\% 
    \multicolumn{1}{c|}{}   & moment to thrust coefficient ($k_{TQ}$)					& $0.02$      			     \\ %  
    \multicolumn{1}{c|}{}   & arm length ($l\,[\rm m]$)                                 & $0.34$     			    \\% 
\bottomrule 
\end{tabular}
\vspace{-0.4cm}
\end{table}

%Compared to the method proposed in \cite{falanga2017aggressive}, we do not need any prior information on the gap size. 
%Besides, the calculation of the Perspective-n-Points (PnP) problem is complex. In order to reduce the error, the method has to use IMU and fuse the result. In this work, we only rely on a binary image and a depth image to calculate the pose, which is easier to implement, less computationally intensive, and more accurate. \textcolor{blue}{(The depth error of D455 camera is less than 2$\%$ within 4m.)}
Compared to the method proposed in \cite{falanga2017aggressive}, our approach does not require any prior information on the gap size. 
Furthermore, our method is simpler to implement, and more computationally efficient, relying only on a binary and a depth image to calculate the pose. %, as it only relies on a binary and a depth image to calculate the pose. 
It is worth noting that the depth error of the D455 camera used in our system is less than 2$\%$ within a range of 4 meters.
In contrast, the approach in \cite{falanga2017aggressive} requires prior knowledge of the gap size, limiting its ability for unknown gap sizes.

%\textbf{\textit{[figure.6] Variations of each reward and curriculum difficulty factor over episodes.}}
\begin{figure}[t] 
\centering
\includegraphics[width=\linewidth]{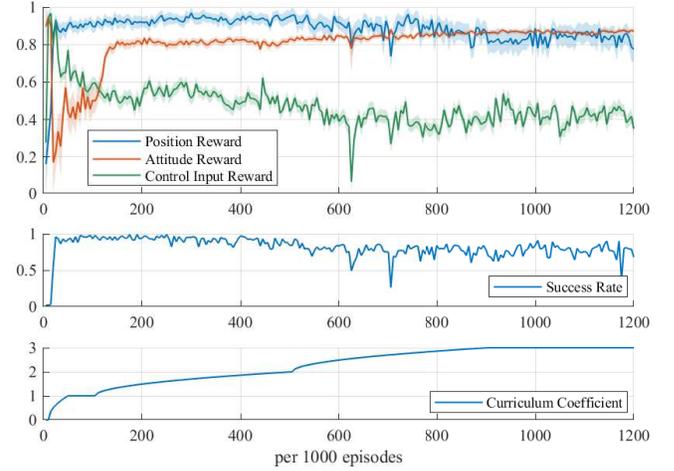}
\caption{
Rewards Learning over Episodes. 
The rewards calculated by \eqref{eq_r_p}-\eqref{eq_r_u} are normalized. 
}
\label{fig_rewards}
\vspace{-0.4cm}
\end{figure}

\vspace{-0.2cm}
\section{Results}
%In this section, the proposed system is evaluated via ...% with respect to the following performance criteria
In this section, we evaluate the proposed system. 
We first transfer the trained policy into a new simulation environment and validate the generalization ability to different domains without more training data. %that our policy can easily be generalized to different domains without more training data. 
Ablation studies are presented to validate the policy algorithm designs for gap attitude variation. %the design choices of the proposed approach.  
% A physical quadrotor is then built to implement the whole system. 
We perform repetitive real-world experiments, demonstrating the effectiveness and robustness of the proposed method. 
At last, we reproduce a traditional gap traversal method \cite{falanga2017aggressive}, and implement experiments to compare the control performances.

\vspace{-0.2cm}
\subsection{Training Configuration}
% \vspace{-0.1cm}
% training configuration
The parameters of the training algorithm defined in Section \ref{sec_learning} are summarized in Table \ref{tab_train_params}. %, together with the SAC hyperparameters based on \cite{haarnoja2018soft}.
%We study the rewards and curriculum difficulty factor over episodes in Figure \ref{fig_rewards}.
The rewards and curriculum difficulty factor over episodes are analyzed in Figure \ref{fig_rewards}. 
As the curriculum difficulty increases, the policy continues to explore and learn, which can lead to local minima. 
This forms reward curves that exhibit generally ascending trends with intermittent spikes. 
% initial
%In the initial stage (the first 100k episodes), the gap is big enough for the vehicle to traverse with any attitude, and the policy learns to fly behind the gap. Thus, the position reward $r_p$ rises quickly in the first 15k episodes. 
Initially (the first 100k episodes), the gap is wide enough for the quadrotor to traverse with any attitude, resulting in a quick rise of the position reward $r_p$ in the first 15k episodes. 
% middle
% As training episodes go on, the vehicle then learns to follow gap's attitude with guidance from attitude reward and also the constraint from the smaller gap. Therefore, the attitude rewards $r_a$ increase in 80k to 180k episodes. 
As the training episodes proceed, the vehicle learns to follow the gap's attitude with guidance from the attitude reward and the constraint from the narrowing gap. 
Hence, the attitude rewards $r_a$ increase during 80k to 180k episodes. 
% last
% As the gap further narrows down, the policy learns to maintain the high reward from completing the task, enabling an accurate whole-body control policy trained out. 
As the gap continues to narrow, the policy learns to maintain a high reward from completing the task, leading to the development of an accurate whole-body control policy. 
% no curriculum
Providing the narrow gap goal from the outset of training, conversely, makes the terminal reward difficult to obtain, posing a challenge for the quadrotor to develop an accurate control policy.

\vspace{-0.3cm}
\subsection{Sim2Real Validation}
% \vspace{-0.1cm}
% simulation env.
Before conducting real-world experiments, simulations in a different environment are implemented to validate the generalization ability of our policy. 
%The simulation platform is a laptop featuring a 3.6GHz 8 core Intel Core i7-7700 CPU and an Intel HD Graphics 630. 
{The software-in-the-loop (SITL) tests are all conducted utilizing Gazebo9 and PX4-Autopilot v1.11, running on a laptop featuring a 3.6GHz 8 core Intel Core i7-7700 CPU and an Intel HD Graphics 630. %} 
The algorithms are implemented in ROS with Ubuntu 18.04.} 
%, which is a powerful 3D simulation environment for autonomous robots that is particularly physically realistic. 
%We utilize the iris quadrotor model to simulate the vehicle with a customized configuration. % mass 0.9,  motorconstant 1.5e-5, diagonal inertia matrix $\bm J = \rm diag(0.019, 0.019, 0.045) kg\cdot m^2$. 
The mass of the drone is set as $0.9{\rm kg}$ with a motor constant of $1.5\times 10^{-5}$. 
Note that the quadrotor dynamics parameters vary from the training environment. 
We only guarantee {enough} thrust-to-weight ratio, which is closely related to the ability of aggressive motion. 
%{
%All the simulations are run on a laptop featuring a 3.6GHz 8 core Intel Core i7-7700 CPU and an Intel HD Graphics 630. 
%The algorithms are implemented in ROS, running on Ubuntu 18.04, and the simulation environment are supported by Gazebo9 and PX4-Autopilot v1.11.
%}

\begin{table*}[]
\centering
\caption{
%{
Evaluation of the Policy and Ablation Study in SITL Simulation Compared with Training Results.
}
%}
\label{tab_success_rate}
\setlength\tabcolsep{4pt}%
\begin{threeparttable}%[b]

\begin{tabular}{rr|ccccccccccccc}
\toprule
    & Methods       & -60$^\circ$ & -50$^\circ$ & -40$^\circ$ & -30$^\circ$ & -20$^\circ$ & -10$^\circ$ 
    & 0$^\circ$     & +10$^\circ$ & +20$^\circ$ & +30$^\circ$ & +40$^\circ$ & +50$^\circ$ & +60$^\circ$ \\
\midrule
    & Training Results  \tnote{1}
    & {\textbf{98.3$\%$}}  & {\textbf{99.1$\%$}}  & {\textbf{99.6$\%$}}  
    & {\textbf{99.4$\%$}}  & {\textbf{99.7$\%$}}  & {\textbf{99.9$\%$}}  
    & {\textbf{99.9$\%$}}  
    & {\textbf{99.9$\%$}}  & {\textbf{99.7$\%$}}  & {\textbf{98.8$\%$}}  
    & {\textbf{98.3$\%$}}  & {\textbf{96.2$\%$}}  & {\textbf{90.5$\%$}}   \\
\midrule
    \multicolumn{1}{c|}{\multirow{3}{*}{Config. 1 }}         & Ours   \tnote{2}
    & {\textbf{83$\%$}}  & {\textbf{92$\%$}}  & {\textbf{94$\%$}}  
    & {\textbf{96$\%$}}  & {\textbf{98$\%$}}  & {\textbf{98$\%$}}  
    & {\textbf{98$\%$}}  
    & {\textbf{99$\%$}}  & {\textbf{97$\%$}}  & {\textbf{97$\%$}}  
    & {\textbf{91$\%$}}  & {\textbf{92$\%$}}  & {\textbf{76$\%$}}   \\
    \multicolumn{1}{c|}{}                   & w/o attitude reward  % \tnote{1} 
    & {47$\%$}    & {62$\%$}    & {73$\%$}    & {82$\%$}    & {95$\%$}    & {96$\%$}
    & {96$\%$}   
    & {98$\%$}    & {94$\%$}    & {82$\%$}    & {80$\%$}    & {69$\%$}    & {43$\%$} 
    \\
    \multicolumn{1}{c|}{}                   & {w/o attitude augment}
    & {48$\%$}    & {49$\%$}    & {76$\%$}    & {80$\%$}    & {92$\%$}    & {88$\%$}
    & {91$\%$}   
    & {90$\%$}    & {84$\%$}    & {83$\%$}    & {70$\%$}    & {69$\%$}    & {36$\%$} 
    \\
\midrule
    \multicolumn{1}{c|}{\multirow{3}{*}{Config. 2}}     & Ours
    & {\textbf{95$\%$}}   & {\textbf{98$\%$}}  & {\textbf{99$\%$}}  
    & {\textbf{100$\%$}}  & {\textbf{100$\%$}}  & {\textbf{100$\%$}}  
    & {\textbf{100$\%$}}  
    & {\textbf{100$\%$}}  & {\textbf{99$\%$}}  & {\textbf{100$\%$}}  
    & {\textbf{99$\%$}}  & {\textbf{98$\%$}}  & {\textbf{88$\%$}}   \\
    \multicolumn{1}{c|}{}                   & w/o attitude reward  
    & {79$\%$}    & {85$\%$}    & {95$\%$}    & {99$\%$}    & {100$\%$}    & {100$\%$}
    & {99$\%$}   
    & {99$\%$}    & {100$\%$}    & {97$\%$}    & {96$\%$}    & {83$\%$}    & {67$\%$} 
    \\
    \multicolumn{1}{c|}{}                   & {w/o attitude augment}
    & {73$\%$}    & {82$\%$}    & {94$\%$}    & {98$\%$}    & {100$\%$}    & {100$\%$}
    & {100$\%$}   
    & {100$\%$}    & {99$\%$}    & {99$\%$}    & {95$\%$}    & {88$\%$}    & {75$\%$} \\
\bottomrule 
\end{tabular}

\begin{tablenotes}
    \item[1] 1000 tests for each case in training environment.
    \item[2] 100 tests for each case in SITL environment.
    \item[3] Configuration 1\&2 are for SITL tests. Configuration 1 follows the training environment with drone size of $0.47{\rm m} \times 0.17{\rm m}$ and gap size of $0.70{\rm m} \times0.30{\rm m}$. Configuration 2 follows the real-world experiments with drone size of  $0.35{\rm m} \times 0.20{\rm m}$ and gap size of $0.70{\rm m} \times0.40{\rm m}$. 
\end{tablenotes}
\end{threeparttable}

\vspace{-0.5cm}
\end{table*}

% action tuning % how we realize sim2real
The control frequencies in the Gazebo simulation and the following real-world experiment are given as $50\rm Hz$, which is different from our training environment, as stated in Section \ref{sec_sub_train_detail}.
%{
Considering the Sim2Real gap in control frequency, quadrotor dynamics, low-level controllers etc., we {tune} the linear mapping scale of the network outputs as $\bm\kappa = [160.0, 160.0, 24.0] $ in SITL and real-world experiments. %}
%The scales given in SITL and real-world experiments are $160.0, 160.0, 24.0$ for $\ddot{\phi}, \ddot{\theta}, \ddot{z}$ respectively. 
%The mapping scale can be found in Figure [].
After that, the policy trained only in our training environment can work well in unknown environments. 
We count the success rate through thousands of SITL tests with respect to different roll angles of the gap, as shown in Table \ref{tab_success_rate}. 
The tests are implemented in two configurations related to the training environment and real-world experiments. 
% In configuration 1, the drone size and gap size follow the training environment as $0.47{\rm m} \times 0.17{\rm m}$ and $0.70{\rm m} \times0.30{\rm m}$. Configuration 2 has the same setup as in the real-world experiments, in which the size of the drone is $0.35{\rm m} \times 0.20{\rm m}$ and the gap is $0.70{\rm m} \times0.40{\rm m}$. 
The high success rates maintained from training environment demonstrate the robustness of the proposed algorithm. 
% Besides success rates, we conduct more tests in SITL to evaluate our policy. 
{
Please refer to the \href{https://github.com/arclab-hku/DRL_Agile_Gap_Traversal/blob/main/SupplementaryMaterial.pdf}{supplementary material} for further evaluation of our policy.}

%% Figure. The quadrotor platform used in the experiments
\begin{figure}[t] 
\centering
\includegraphics[width=0.95\linewidth]{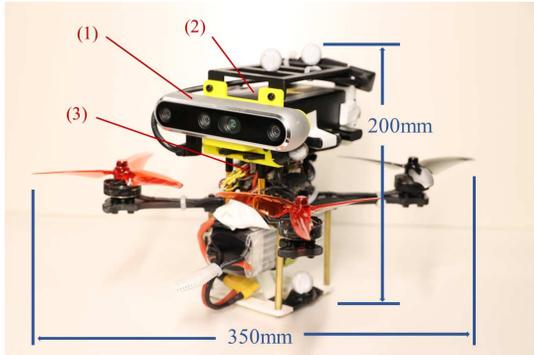} 
\caption{Our quadrotor platform for real-world validation.
(1) Intel D455i depth camera. 
(2) DJI Manifold 2-C onboard computer.
(3) Pixracer. 
}
\label{fig_drone_platform}
\vspace{-0.4cm}
\end{figure}

\vspace{-0.3cm}
\subsection{Ablation Study}
We perform ablation studies to validate the designs of the proposed approach. 
%To handle the problem of the gap attitude variation, we design an attitude augment on policy inputs and the attitude reward to enhance the robustness of our policy. 
%Therefore, we focus on the effect of the attitude augment and the attitude reward. 
Specifically, we focus on the effect of attitude augmentation (on network inputs) and attitude reward, which are designed to handle the variation of gap attitude. 
{
We replace the attitude augmentation with a plain attitude input or ablate the attitude reward in training and deploy the resulting policy into SITL tests.} 
%We ablate the attitude augmentation or attitude reward in training and deploy the resulting policy into SITL tests. % for different gap roll angles. 
The success rates are counted and summarized in Table \ref{tab_success_rate}. 

For the ablation study on attitude reward, the results show that success rate is not affected significantly when the roll angle is small, i.e., $|\phi_g|\le20^\circ$. 
However, when the task gets more complicated, e.g., increasing the attitude of gap or decreasing the size ratio of gap to drone, training without attitude reward is insufficient to achieve the traversal. 
Meanwhile, the attitude augmentation on network inputs increases the success rate and robustness of the policy for most situations. 
The results indicate the necessity of attitude augmentation for our policy training. 
% Furthermore, the results indicate that training with attitude augmentation increases the robustness 

\begin{table}[t]
\centering
\caption{Gap Detection Error Statistics}
\label{tab_gap_detect_error}
\begin{tabular}{c|ccc|ccc}
    \toprule
                & \multicolumn{3}{c|}{Position error $[m]$}     & \multicolumn{3}{c}{Orientation error $[^\circ]$} \\ 
    \midrule
                & $\Delta x$    & $\Delta y$        & $\Delta z$    & $\Delta\phi$      & $\Delta\theta$  & $\Delta\psi$\\ 
    \midrule
    $\mu$    	& {0.058}    & {0.045}    & {0.022}    
                & {2.907}    & {4.494}    & {2.240}    \\
    $\sigma$ 	& {0.006}    & {0.007}    & {0.006}    
                & {1.759}    & {1.914}    & {1.578}    \\
    \midrule
    {$10\%$ CI}      & {0.050}    & {0.037}    & {0.015}    
                                    & {0.670}    & {1.682}    & {0.180}    \\
    {$95\%$ CI}      & {0.069}    & {0.055}    & {0.032}   
                                    & {6.087}    & {6.869}    & {4.427}    \\
    \bottomrule
\end{tabular}
\vspace{-0.4cm}
\end{table}

%% table: traversing error
\begin{table}[t]
\centering
\caption{Pose Error Statistics at Traversal Point.}
\label{tab_traverse_error}
\begin{tabular}{c|cc|cc}
    \toprule
    & \multicolumn{2}{c|}{Position error $[m]$}     & \multicolumn{2}{c}{Orientation error $[^\circ]$} \\ 
    \midrule
                  & $\Delta y$    & $\Delta z$    & $\Delta\phi$  & $\Delta\theta$ \\  
    \midrule
    $\mu$    	& {0.065}    & {0.033}    & {5.297}    & {6.749}    \\
    $\sigma$ 	& {0.047}    & {0.036}    & {4.494}    & {6.745}    \\
    \midrule
    {$10\%$ CI}      & {0.004}    & {0.005}    & {0.6207}    & {0.9167}    \\
    {$95\%$ CI}      & {0.145}    & {0.093}    & {12.865}    & {19.538}    \\
    \bottomrule
\end{tabular}
\vspace{-0.5cm}
\end{table}

\begin{table*}[]
\centering
\caption{
{
Success Rate in Real-World Experiments. 
}
}
\label{tab_success_rate_expr}
\setlength\tabcolsep{4pt}%
\begin{threeparttable}%[b]

\begin{tabular}{r|ccccccccccc}
\toprule
    Gap Roll Range$[^\circ]$    & [-60, -50)    & [-50, -40)    & [-40, -30)    & [-30, -20)    & [-20, -10)
                & [-10, 10]     & (10, 20]      & (20, 30]      & (30, 40]      & (40, 50]      & (50, 60]        \\
\midrule
    Experiment Results  \tnote{1}   & 100.0$\%$     & 87.5$\%$      & 85.7$\%$      & 88.9$\%$      & 100.0$\%$
                                    & 100.0$\%$     & 100.0$\%$     & 100.0$\%$     & 72.7$\%$      & 71.4$\%$      & 80.0$\%$  \\
\bottomrule 
\end{tabular}

\begin{tablenotes}
    \item[1] At least 5 traversal flights for each case.
\end{tablenotes}
\end{threeparttable}

\vspace{-0.5cm}
\end{table*}

%% figure: drone states during traversing
\begin{figure}[t] 
\centering
\includegraphics[width=\linewidth]{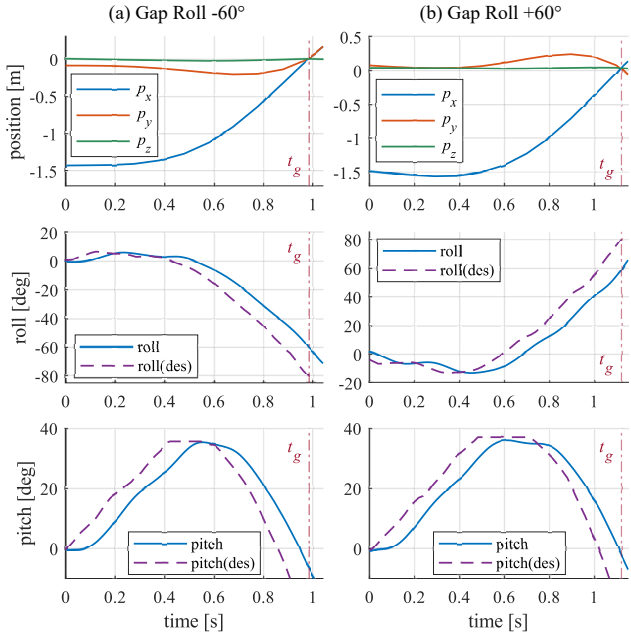}
\caption{
{
Drone states over time during narrow gap traversal. 
Each column depicts the results of an experiment performed under distinct gap attitudes: (a) $-60^\circ$ roll and (b) $+60^\circ$ roll. 
The quadrotor reaches the center of the gap at $t=t_g$. 
}
}
\label{fig_dronestates}
\vspace{-0.5cm}
\end{figure}

\vspace{-0.2cm}
\subsection{Real-World Experiments} 
\subsubsection{Experiment Setup}
\label{sec_expr_setup}

% \subsubsection{}
% hardware setup
We validate our proposed system in the real world. 
Figure \ref{fig_drone_platform} shows our quadrotor platform used in the experiments. 
The target gap is detected by an Intel D455i depth camera. 
%Our quadrotor is equipped with an Intel D455i depth camera which detects the target gap. 
The gap observation algorithm and control policy runs on a DJI Manifold 2-C computer, sending low-level attitude and altitude control commands to a Pixracer. % PX4 Autopilot. 
% that contains an IMU and a micro controller on which the low-level controller runs. 
All real-world experiments are conducted indoors with a motion capture system, which facilitates state observation of the vehicle. 
% Thus, the state of the drone is computed by fusing message from the motion capture system and the onboard IMU. 

% drone configuration params
The overall weight of our quadrotor is $1.1 \rm kg$, with a thrust-to-weight ratio of $3.5$. 
The arm length of the quadrotor is $22\rm cm$, and the overall dimension is $35 \rm cm \times 20\rm cm $ (the largest length measured between propeller tips), while the size of the gap used in experiments is $70 \rm cm \times 40\rm cm $. 
When the vehicle is at the center of the gap, the long and short sides tolerances are only $17.5\rm cm$ and $10\rm cm$, respectively. 
%A remote control module is configured to rotate the gap to arbitrary roll angle. 
%To demonstrate the effectiveness of our method, 
In our experiments, the drone aims to fly through a variable-angle narrow gap back and forth.

\subsubsection{Experiment Results}
%\subsection{Real-World Experiment Results} 
\label{sec_real_world_expr}
We design groups of experiments to demonstrate the robustness of the proposed gap detection algorithm as well as the control policy. 

The accuracy of our onboard sensing method is first evaluated. 
We detect the narrow gap placed in different poses and compare the results with ground truth data from a motion capture system. 
The statistics of the measurement error are shown in Table \ref{tab_gap_detect_error}. 

We then implement repetitive experiments to evaluate the whole system, where the quadrotor is required to traverse multiple gaps with different roll angles. 
Overall, we ran {87} traversals with the roll angle ranging from $-60^\circ$ to {$+60^\circ$}, achieving a remarkable success rate of {87.36$\%$}. 
{Success rates with respect to different angle ranges are calculated in Table \ref{tab_success_rate_expr}.} 
Figure \ref{fig_dronestates} shows the traversal motion with estimated position and orientation over time in two representative experiments. 
It can be observed that the drone orientations are planned precisely by the policy, resulting in an almost perfect posture of the vehicle when it reaches the gap plane. 
Specifically, at time $t_g$, the roll is close to the gap, and the pitch reduces to zero, while the position is close to the gap center. 
Table \ref{tab_traverse_error} reports the statistics of the pose errors at time $t=t_g$, measured as a distance between drone posture and the gap. 
The errors include control errors introduced by control policy and detection errors introduced by gap detection algorithm. 
The statistics include both successful and unsuccessful experiments. 
Compared to the traversal error statistics result using traditional optimization-based method in \cite{falanga2017aggressive}, our framework achieves comparable results, demonstrating the robustness and the potential of exploiting the quadrotor's agility of learning-based methods.

%We further considered a scenario for the quadrotor to fly back and forth through a gap with an increasing inclined angle. 
%The experiment results are provided in Figure \ref{fig_traversing}(b). Furthermore, we have tested our algorithms in the presence of varying illuminations (turn off the lights). We refer the reader to the accompanying video for more experiment details at {https://youtu.be/HUTWBclayT8}. 

\vspace{-0.2cm}
\subsection{ {Comparative Study with Traditional Method}} % with
\label{sec_compare_study}

\begin{figure}[t] 
\vspace{-0.4cm}
\centering
\includegraphics[width=\linewidth]{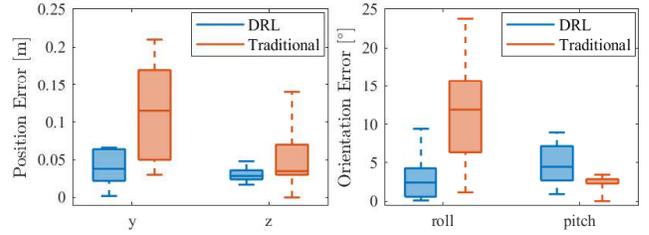}
\caption{{
Comparison of Traversal State Error.
}}
\label{fig_cmp_error}
\vspace{-0.3cm}
\end{figure}

\begin{figure}[t] 
\centering
\includegraphics[width=\linewidth]{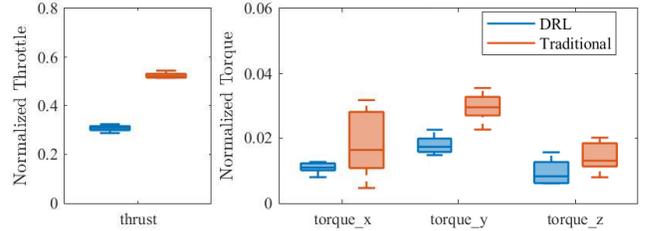}
\caption{{
Comparison of Actuator Control Effort.
}}
\label{fig_cmp_actuator}
\vspace{-0.5cm}
\end{figure}

{
We compare the proposed traversal policy with a traditional method in \cite{falanga2017aggressive}, which designed a two-stage traversal trajectory based on the differential flatness property of quadrotors. 
%typically decoupled trajectory planning and control, designed an optimization-based approach and traverse trajectory. 
%The method designed a optimization-based two-stage trajectory. 
We implement the trajectory planning method and control algorithm used in \cite{falanga2017aggressive} on the same platform specified in Section \ref{sec_expr_setup}.
%, and compare it with our traversal policy in real-world experiments. 
% experiments
%Note that we focus on the performance comparison of traditional method and learning-based method for the gap traversal flight. Therefore, the gap pose is detected by the motion capture system in this section. 
As our main focus was on the control performance comparison, we employed a motion capture system to accurately detect the gap pose. 
As the maximum tilt angle of the gap is $45^\circ$ in the experiments of \cite{falanga2017aggressive}, 
we performed tests in five different scenarios with gap roll angles of $0^\circ$, $\pm20^\circ$, $\pm45^\circ$. Each scenario was repeated twice. 
%we test the flight performances of two methods for five different inclined orientations at $0^\circ$, $\pm20^\circ$, $\pm45^\circ$. Each case is tested twice. 
%We statistic and compare the traversal state error and the actuator control efforts for each method, as shown 
We compared the traversal state error and actuator control efforts of each method, and the results are presented in Figures \ref{fig_cmp_error} and \ref{fig_cmp_actuator} respectively. 
% We statistic and compare the traversal state error for each method, as shown in Figure \ref{fig_cmp_error}. 
}

{
The computation time is compared in Python. 
For each control step, the traditional method takes $1.025 {\rm ms}$ to re-plan the trajectory and compute control commands, while the proposed method only requires $0.615 {\rm ms}$ to generate and process network outputs. 
Further discussions of the comparison results are presented in Section \ref{discuss_traverse}. 
}

\vspace{-0.2cm}
\section{{Discussion}}
{
In this section, we discuss our system and provide more insights into the proposed methods.
}

\vspace{-0.2cm}
% 1. control performance compared between traditional and drl
\subsection{{Gap Traversal Control Policy}}      \label{discuss_traverse}
{
Traditional quadrotor agile flight methods typically decouple trajectory planning and control. 
%The conventional methods  usually follow the pipeline of planning a trajectory and then tracking the trajectory by the controller
The performance and success rate depend highly on both the quality of the planned trajectory and the controller. 
Towards narrow gap traversal flight, the conventional approaches focus on planning dynamically feasible trajectories by exploiting differential flatness of the quadrotor \cite{loianno2016estimation, falanga2017aggressive}. 
%An efficient controller based on accurate system dynamics is required to correct the drone state at a high rate. 
%These trajectories are inherently smooth. Hence, they cannot represent the rapid state or input changes in a reasonable order, and only reach the input limits for an infinitesimal short duration \cite{penicka2022minimum}. 
% Although the algorithms have been carefully designed and tuned, a tiny low-level control delay in the real-world implementations could lead to certain control errors in the aggressive motion (i.e., linear velocity up to $3 \rm{ m/s}$, angular velocity up to  $4 \rm{rad/s}$), as shown in Figure \ref{fig_cmp_error}.
It is important to note that despite careful design and tuning of the algorithms, the minimal low-level control delays in the real-world implementations will lead to certain control errors during aggressive motions (i.e., linear velocity up to $3 \rm{ m/s}$, angular velocity up to $4 \rm{rad/s}$), as shown in Figure \ref{fig_cmp_error}. 
When the motion is relatively moderate (e.g., the pitch angle), the traditional method can perform better. 
% A better control performance could exhibit when the motion is relatively moderate (e.g. the pitch control in Figure \ref{fig_cmp_error}). 
In contrast, our learning-based method provides an end-to-end policy 
% that learn and compensate for the features of the control response during policy training.
that learns and adapts to control response features during training. 
%\textcolor{magenta}{
% Therefore, it requires less effort in controller design and tuning but can also achieve better performance for aggressive maneuvers. 
This eliminates the need for extensive controller design and tuning while still achieving better performance in aggressive maneuvers. 
% has the potential to achieve better performance. %} 
%are able to learn and compensate for the features of the control response during policy training, therefore requiring less effort on controller tuning. 
Moreover, our method requires fewer control efforts to accomplish the task, as indicated by the comparison result in Figure \ref{fig_cmp_actuator}. 
This demonstrates the effectiveness of the control input penalty \eqref{eq_r_u} and the exploration capabilities of our proposed method.
% Moreover, the comparison result of the actuator control effort shows that our policy is able to accomplish the task with fewer control efforts, indicating the effectiveness of the penalty on control input \eqref{eq_r_u} and the exploration ability of the proposed method. 
Although the traditional method has derived closed-form solutions for re-planning, the proposed method with a lightweight end-to-end policy exhibits superior performance in computation time. 
One limitation of the proposed learning-based method is that it only has soft constraints by giving rewards. 
%, which would raise concerns in safety-critical robotics applications.
%, but lacks of hard constraints. 
}

\vspace{-0.2cm}
% 2. Scalability and generalizability
\subsection{{Scalability and Generalizability}}
{
We further explore the scalability and generalizability of the system through complementary experiments. 
Regarding gap detection, we successfully test our algorithm under varying illuminations and changing the gap size in the real-world experiments, respectively. 
However, the proposed gap detection method is limited to the marked rectangular frame. 
A structure-less gap detection method could be considered in future work refer to \cite{sanket2018gapflyt}. 
For the control policy, we successfully test our algorithm with different quadrotor dynamics in SITL, while only requiring enough thrust-to-weight ratio (more than 2.5 for up to $60^\circ$ maneuvers). 
Furthermore, to test the ability of the whole proposed system, we considered a scenario for the quadrotor to fly back and forth through a gap with an increasing inclined angle. 
The experiment results are provided in Figure \ref{fig_traversing}(b). 
We refer the reader to the accompanying video for more experiment details at \href{https://youtu.be/HUTWBclayT8}{https://youtu.be/06F6YDsypPQ}.
}

\vspace{-0.2cm}

\section{Conclusion}
This letter presented a learning-based system for a quadrotor to fly through an unknown tilted narrow gap. 
%We propose a reinforcement learning algorithm to train a traversal flight control policy. 
Compared to our previous work, the training algorithm incorporated an input augmentation and a carefully designed reward function to handle variation in gap attitude. 
%We demonstrate the effectiveness and robustness of the proposed training algorithm with thousands of Sim2Real tests and ablation studies. 
%We further introduced an onboard sensing method into our system and performed autonomous gap detection. %, so that no prior knowledge of the gap is required
Additionally, an onboard sensing method is introduced for autonomous gap detection, eliminating the need for prior environmental knowledge. 
%We validated our end-to-end system in real-world experiments, achieving a success rate of ...
The end-to-end system is validated through real-world experiments, achieving a success rate of 
{87.36$\%$} in {87} traversals.
%To the best of our knowledge, this is the first work that performs the learning-based traversal flight through a variable-tilted narrow gap in the real world, without prior knowledge of the environment. 
To the best of our knowledge, this is the first work that performs the learning-based traversal of variable-tilted narrow gaps in the real world without prior knowledge of the environment. 

%% further work
% One limitation of this study is that the orientations are represented by Euler angles, which could introduce singularity issues for some extreme states. 
One limitation of this study is that using Euler angles to represent orientations may introduce singularity issues for some extreme states. 
%Consequently, the policy cannot handle a full-state \textit{SE(3)} problem because of the singularity issues raised by Euler angles. 
%Future work would address a full-state \textit{SE(3)} flight using rotation matrix or quaternion representations. 
Future work will explore full-state \textit{SE(3)} flight using rotation matrix or quaternion representations.

\vspace{-0.2cm}
\section*{acknowledgement}
The authors gratefully thank Yunfan Ren and Yixi Cai for their help in picture-making and helpful discussions. 

% \appendix

% \bibliographystyle{apalike-etal-in-italics}

\vspace{-0.2cm}
\bibliographystyle{IEEEtran}
\bibliography{root.bib}

\end{document}